\documentclass{article} 
\usepackage{iclr2025_conference,times}
\usepackage{natbib}

\usepackage{amsmath,amsfonts,bm}

\usepackage{amsthm}

\theoremstyle{definition}
\newtheorem{definition}{Definition}[section]

\def\eqref#1{equation~\ref{#1}}

\def\1{\bm{1}}

\DeclareMathAlphabet{\mathsfit}{\encodingdefault}{\sfdefault}{m}{sl}
\SetMathAlphabet{\mathsfit}{bold}{\encodingdefault}{\sfdefault}{bx}{n}

\usepackage{amsthm}
\usepackage{hyperref}
\usepackage{url}
\usepackage{graphicx}
\usepackage{booktabs}
\usepackage{subcaption}
\usepackage{enumitem}

\title{Looking Beyond the Top-1: \\Transformers Determine Top Tokens in Order}

\author{
    \begin{minipage}[t]{0.48\textwidth}
    \centering
    Daria Lioubashevski\thanks{Corresponding author. Email: daria.lioubashevski@mail.huji.ac.il} \\
    \textnormal{The Hebrew University of Jerusalem}
    \end{minipage}
    \hfill
    \begin{minipage}[t]{0.48\textwidth}
    \centering
    Tomer Schlank \\
    \textnormal{The University of Chicago}
    \end{minipage} \\
    \\ 
    \begin{minipage}[t]{0.48\textwidth}
    \centering
    \textbf{Gabriel Stanovsky} \\
    The Hebrew University of Jerusalem
    \end{minipage}
    \hfill
    \begin{minipage}[t]{0.48\textwidth}
    \centering
    \textbf{Ariel Goldstein} \\
    The Hebrew University of Jerusalem
    \end{minipage}
}

\makeatletter
\iclrfinaltrue
\makeatother

\begin{document}
\maketitle
\fancyhead[LO]{Preprint. Under review.}

\begin{abstract}
Understanding the inner workings of Transformers is crucial for achieving more accurate and efficient predictions. In this work, we analyze the computation performed by Transformers in the layers after the top-1 prediction has become fixed, which has been previously referred to as the ``saturation event''. We expand the concept of saturation events for top-$k$ tokens, demonstrating that similar saturation events occur across language, vision, and speech models. We find that these saturation events happen {\emph{in order}} of the corresponding tokens' ranking, i.e., the model first decides on the top ranking token, then the second highest ranking token, and so on. This phenomenon seems intrinsic to the Transformer architecture, occurring across different architectural variants (decoder-only, encoder-only, and to a lesser extent full-Transformer),  and even in untrained Transformers. We propose an underlying mechanism of task transition for this sequential saturation, where task $k$ corresponds to predicting the $k$-th most probable token, and the saturation events are in fact discrete transitions between the tasks. In support of this we show that it is possible to predict the current task from hidden layer embedding. Furthermore, using an intervention method we demonstrate that we can {\emph{cause}} the model to switch from one task to the next. Finally, leveraging our findings, we introduce a novel token-level early-exit strategy, which surpasses existing methods in balancing performance and efficiency.
\end{abstract}

\section{Introduction}
In recent years, Transformer-based models \citep{vaswani2017attention} have achieved state-of-the-art performance in various tasks across multiple modalities, including text generation, image classification, and automatic speech recognition \citep{zhang2023google,openai2024gpt4technicalreport}. This has lead to a growing interest in model interpretability, which tries to explain the internal processes that give rise to these remarkable capabilities. In the language domain, investigation into the way the model's predictions are constructed has led to the discovery of \emph{saturation events}, where the model’s top-1 prediction is determined in an early layer and remains fixed in  subsequent layers~\citep{geva2022transformer}. 

\begin{figure}[t]
\begin{center}
\includegraphics[width=0.8\linewidth]{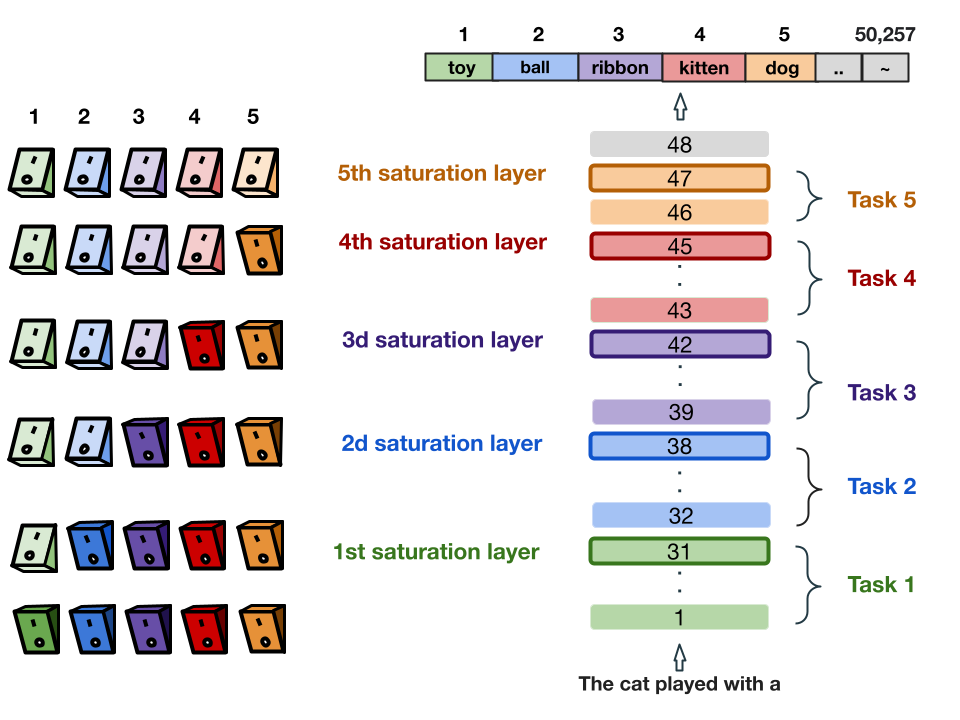} 
\end{center}
\caption{An illustration of the proposed task-transition mechanism wherein the layers of the Transformer perform a changing number of tasks in order, so that task \(i\) is determining the \(i\)-th token in the final ranking, and the transition between task \(i\) and task \(i+1\) occurs at the corresponding \(i\)-th saturation layer. The transition is akin to a switch being flipped ``on" and staying ``on" for the remaining layers representing the \(i\)-th token being fixed from this point onward.}
\label{fig:task_transition}
\end{figure}

In this work, we address the following question – \emph{what computation is the Transformer model performing \textit{after} the saturation event?} 
Taking inspiration from \citet{frydenlund2022language}, we treat the model’s output as a ranking over the labels instead of a probability distribution. Using the logit lens~\citep{nostalgebraist2020gpt}, we project the hidden states of intermediate 
layers onto the vocabulary space to extract ranking over tokens and analyze the changes over the layers. For the first time, we show that in a decoder-only text Transformer (GPT2-XL; \citealp{radford2019language}) saturation events also take place for the top ranking tokens beyond the top-1 (2nd, 3rd, 4th, etc.).
Surprisingly, we find that they happen \emph{in order} of their ranking, i.e. the second-ranking token is determined only after the first-ranking token, and so forth (see in Figure~\ref{fig:task_transition}).
We then generalize the results across different modalities and Transformer variants, 
including pretrained Transformers for both vision (encoder-only ViT-L/16; \citealp{dosovitskiy2020image}) and speech (encoder-decoder Whisper-large; \citealp{radford2023robust}). Next, we demonstrate that sequential saturation seems intrinsic to the Transformer architecture, occurring even in an \emph{untrained} randomly initialized model (GPT2-XL). 

We propose that this phenomenon is due to a discrete \emph{task-transition} mechanism, where each task \(i\) corresponds to the model determining the \(i\)-th token in the final ranking, and the transition between one task and the next happens at the corresponding saturation layer. Furthermore, we find that the task information is encoded in the layer embeddings and that at each saturation layer, a discrete ``switch" is flipped, signaling that the relevant token has been determined, causing the model to move on to the next task while keeping this token fixed in subsequent layers. To support this, we show that it is possible to predict the task index from the layer embeddings using a simple logistic regression classifier, and that we can cause the model to transition from the first to the second task by ``injecting'' embeddings from either the top-1 saturation layer or of one of the subsequent layers.

Finally, we show that our findings lend themselves to practical applications in improving both model efficiency and accuracy. Based on this new understanding of the Transformer, we define a new early-exit decision strategy for text generation. Early exiting is a technique where the model can make a prediction and terminate the computation before reaching the final layer, thus improving efficiency~\citep{schwartz2020right}. In our method, the early-exit layer is the first one predicted to belong to task 2, presuming that after the transition from task 1 to task 2, the top-1 token represents the model’s final prediction. We show that this strategy outperforms existing token-level measures, such as softmax-response and hidden-state saturation~\citep{schuster2022confident}. In addition, we show that we can use task information to achieve more accurate language models, by demonstrating that in cases where the top-1 prediction is incorrect, the second highest ranking token represents a much more accurate prediction when it reaches saturation than when it does not.   

Our main contributions are: 
\begin{itemize}[left=0pt]
\item We find that Transformers tend to decide their top ranking tokens in order, so that the top ranking token is fixed first, then the second-ranking token at a later layer and so on. We show that this occurs across various modalities and variants of the Transformer architecture, and even in untrained randomly initialized models.
\item We show that sequential saturation can be explained with a discrete \emph{task-transition} mechanism,  encoded in the representation of hidden layers where each task corresponds to determining the next ranking token. We empirically show that it is possible to predict the task index only from internal activations, and that we can cause the model to switch from one task to the next via an intervention procedure.
\item We show that these observations can be leveraged to achieve better downstream performance in early exiting and language modeling, in terms of both accuracy and efficiency.
\end{itemize}
The code for our experiments is available
at \href{https://github.com/daria-lioubashevski/beyond_top1}{https://github.com/daria-lioubashevski/beyond\_top1}.

\section{Experiments}
\begin{figure}[tb!]
    \begin{center}
    \includegraphics[width=0.9\linewidth]{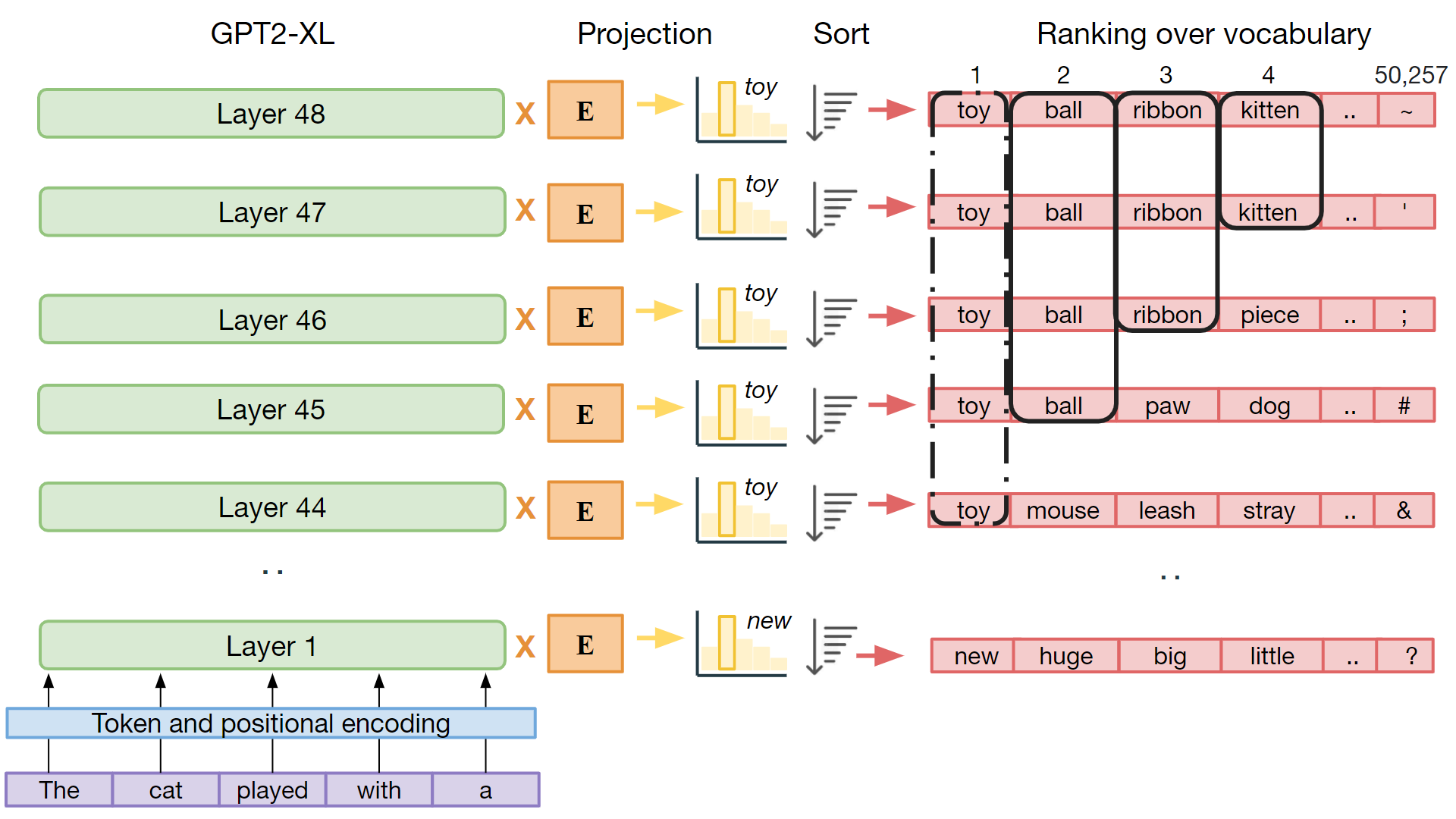} 
\end{center}
\caption{\label{fig:saturation} Schematic of our framework and visualization of the ordered saturation of the top-k tokens on GPT2-XL. The hidden states from each layer are projected onto the vocabulary space using the unembedding matrix \(E\), then sorted in descending order and treated as rankings. The saturation effect is marked separately for each token in the top-4 of the final ranking, emphasizing the fact that the 2nd token saturates \textit{after} the 1st token, the 3d token saturates \textit{after} the 2nd token and so on. The dashed line represents the previously established saturation event of the top-1 token.}
\label {fig:framework}
\end{figure}

In this section, we formulate two experiments to understand what computation the Transformer is performing in the layers after the top-1 saturation event. To achieve this, we first extend the formal definition of top-1 saturation to account for arbitrary i-th ranking token saturation (Section~\ref{sec:defs}). Then, in Section~\ref{sec:order-exp}, we leverage this definition to develop a metric capturing the extent to which the top tokens are saturated in order. In Section~\ref{sec:transition-exp}, we describe a probing method measuring whether it is possible to determine the rank of the token currently being considered by the model solely from the intermediate representation of the layer, without any additional context.

\subsection{Defining Saturation Layers}
\label{sec:defs}

\begin{definition}[1st Saturation Layer; \citealp{geva2022transformer}]
\label{def:top-1-saturation}
 The saturation event occurs at layer $l$ (from here on referred to as the ``1st saturation layer") for index $i$ in the input if the top-1 prediction of the model remains constant in all subsequent layers after $l$. Formally, given a model with $N$ layers, a saturation event occurs at layer \(l \leq N-1 \) if for all layers \(l'\) s.t. \(l < l' \leq N\) the top-1 token in the ranking induced by that layer remains unchanged.
For example, the saturation event marked with a dashed line in Figure~\ref{fig:saturation} occurs on layer 44, since in subsequent layers the top predicted token (``toy'') remains constant.
\end{definition}

\begin{definition}[$k$-th Saturation Layer]
\label{def:top-k-saturation}
Here, we are interested in examining model behavior beyond the determination of the top ranking token and so naturally extend the definition of top-1 Saturation (Definition~\ref{def:top-1-saturation})  to capture the layer at which the $k$-th top token is determined and remains fixed.
Formally, the saturation event for the \(k\)-th top token at index $i$ in the input occurs at layer \(l^k \leq N-1\)  (from here on referred to as the `` k-th saturation layer”) if for all following layers \(l'\) s.t. \(l^k< l' \leq N\) the token in position \(k\) in the ranking induced by that layer remains unchanged. We note that the saturation event defined in \citep{geva2022transformer} happens at $l^1$.
For example, in Figure~\ref{fig:saturation}, $l^1 = 44$ as that is where the top token (``toy'') is determined; $l^2 = 45$, since the second-most probable token (``ball'') is determined at layer 45; and similarly $l^3 = 46$, since the third-most probable token (``ribbon'') is determined at layer 46, etc.\footnote{In all of our experiments, we only consider tokens in the input where the 1st saturation layer satisfies that \(l^1 \leq 0.85 \cdot N\), to ensure that there are enough layers after it for meaningful analysis.}
\end{definition}

\subsection{Examining the Order of Saturation Layers}
\label{sec:order-exp}

We investigate whether the saturation layers of the top-k tokens $l^1, l^2, \ldots l^k$ are arranged in order, i.e., whether the saturation of the first token happens before the saturation of the second token, the saturation of the second token happens before the saturation of the third token and so on. To this end, for each token in the input we first calculate these saturation layers for $k=1, \ldots, 5$ according to Definition~\ref{def:top-k-saturation} and then for each $k$ compute the rank of the saturation layer of the $k$-th top token. We use $k=5$ to ensure consistency across models, as it is the highest value of $k$ where the $k$-th token reaches saturation in at least $5\%$ of input tokens for all the models analyzed.
Following our example from Figure~\ref{fig:saturation}, we have $l^1 = 44, l^2 = 45, l^3 = 46, l^4 = 47$,\footnote{$l^5$ is ill-defined in this case as the 5-th token doesn't reach saturation before the last layer.} and their ranking is $[1,2,3,4]$, since $l^1 < l^2 < l^3 < l^4$. 
If the tokens reach saturation in order of ranking, as they do in this case, we would expect the average rank of the saturation layers to increase monotonically with $k$.

\subsection{Probing for Task Transition}
\label{sec:transition-exp}
We argue that the mechanism underlying the saturation of the top-k tokens in order is one of task transition, such that determining the identity of each token in the final ranking is a separate task, and the model performs them sequentially: first determining the identity of 1st token, then the identity of the 2nd token, and so on, and that the transition from one task to the next occurs at the corresponding saturation layer. Additionally, we claim that the specific task number can be inferred from the model embedding at each layer, and that this information is independent of the context or the specific token predicted by the model. 

To test this hypothesis, we perform a type of probing in which we train a simple one-versus-all multi-class logistic regression classifier to predict the number of the task the model is “working on” from the hidden state embeddings of the model.  We collect the data for training by extracting the model’s hidden states during inference and categorize them into 5 classes according to the saturation layers of the top-5 tokens for each instance. This means that for a given input, embeddings from layers up to (and including) the 1st saturation layer  are classified as belonging to task 1, embeddings from layers from the next layer until the 2nd saturation layer are classified as belonging to task 2, and so forth, for tasks 1 through 5. For example, in the case of the token "a" as depicted in Figure~\ref{fig:saturation}, the embeddings from layers 1 through 44 would be classified as belonging to task 1; the embedding of layer 45 would be classified as belonging to task 2; the embedding of layer 46 would be classified as belonging to task 3; the embedding of layer 47 would be classified as belonging to task 4; and as the model reaches the last layer directly afterward there would be no embedding classified as belonging to task 5.

We balance the training data so that embeddings from all layers are represented equally in each class. To show that the task number is encoded in the model’s embeddings and is not an artifact of the classifier’s weights, we construct a control setting where, for each class, we generate random vectors with the same dimension, drawn from a normal distribution with the mean and variance of the layer embeddings in that class.  

\section{Results}
To show the robustness of our findings, we test pretrained Transformer models on corresponding datasets from three modalities: text, vision and speech.

\begin{figure}[t]
\centering
\begin{subfigure}[b]{0.35\textwidth}
    \centering
    \includegraphics[width=\textwidth]{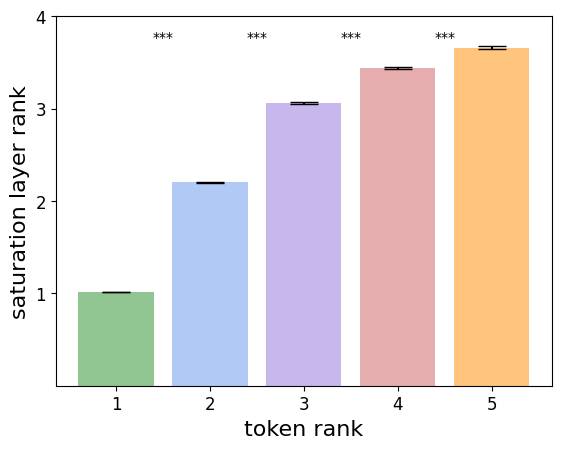}
    \caption{Pretrained GPT2-XL}
    \label{fig:monton_gpt2}
\end{subfigure}
\quad 
\begin{subfigure}[b]{0.35\textwidth}
    \centering
    \includegraphics[width=\textwidth]{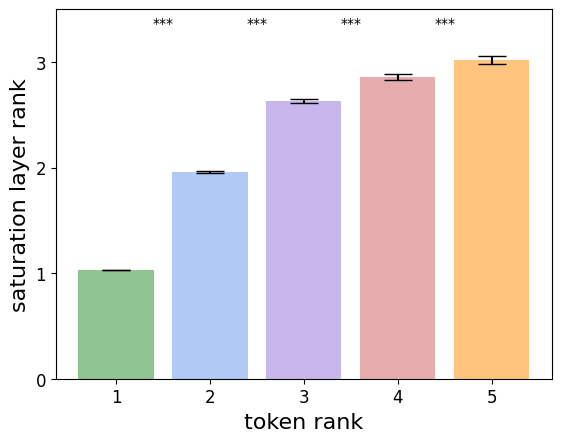}
    \caption{ViT-L/16}
    \label{fig:monton_vit}
\end{subfigure}
\vskip\baselineskip
\begin{subfigure}[b]{0.35\textwidth}
    \centering
    \includegraphics[width=\textwidth]{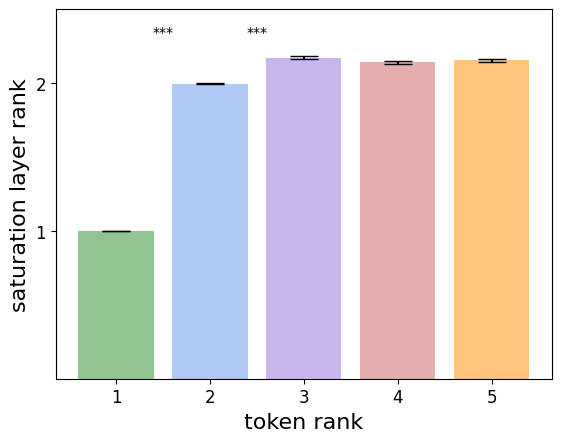}
    \caption{Whisper-large}
    \label{fig:monton_whisper}
\end{subfigure}
\quad 
\begin{subfigure}[b]{0.35\textwidth}
    \centering
    \includegraphics[width=\textwidth]{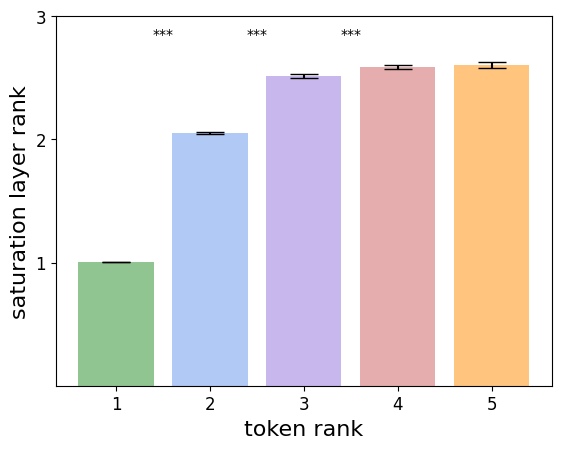}
    \caption{Randomly initialized GPT2-XL}
    \label{fig:monton_random}
\end{subfigure}
\caption{Average rank of the $k$-th saturation layer among the saturation layers for k=1,..,5 with standard error bars. Asterisks indicate statistically significant differences between consecutive token ranks (*** represents $ p < 0.001$), based on an independent samples t-test.}
\label{fig:monton}
\end{figure}

\subsection{Text Transformer}
We first conduct our experiments on a pretrained GPT2-XL model, an auto-regressive decoder-only LLM, using 60K tokens taken from 100 randomly sampled texts from CNN/DM dataset \citep{hermann2015teaching}. 

\textbf{Tokens reach saturation in order of ranking.} Figure \ref{fig:monton_gpt2} shows the average rank of the k-th saturation layer for each $k$. This value increases monotonically with $k$, and the difference between each two consecutive token ranks is statistically significant with $ p < 0.001$ based on a pairwise independent samples t-test. This supports our claim that saturation events happen sequentially according to token ranking in this LLM.
To statistically validate this phenomenon we use a stricter version of Kendall’s {$\tau$} coefficient, where we also consider ties as disagreements  (see Appendix \ref{sec:apx_kendal_tau} for details and mathematical formulation). This is done to discount cases where two or more tokens reach saturation at the same layer. The coefficient takes values in the range \([-1,1] \) where values close to 1 indicate strong agreement, and values close to -1 indicate strong disagreement between the rankings. To check whether the sequence of saturation layers of the top-\(k\) tokens \((l^1, .., l^k)\) is strictly increasing, we use that sequence as one ranking, and the sequence \((1,2,..,k) \) as the other. \(k\) is set independently for each token in the input to be the largest token index such that this token's reaches saturation by our definition i.e. \(l^k<N\). The average $\tau$ coefficient indicates moderate agreement between the rankings, which is larger than all values over 1K permutations, where the saturation layers sequence were randomly shuffled for each instance, resulting in $p<0.001$. 

\textbf{Task number can be predicted from model embeddings.} We split the data into train and test using 5-fold cross validation, and report the mean and standard error of the accuracy. Table \ref{tab:probing_result} shows that the logistic regression classifier trained on embeddings extracted from pretrained GPT2-XL model achieves very high accuracy, while the classifier trained on the random embeddings in the control setting performs approximately at chance level (see Appendix \ref{sec:per_class_probing_metrics} for accuracy and ROC-AUC scores per class).
From this we conclude that the representations of the hidden layers across examples encode task specific information and that the saturation layers as we defined them are the points of transition between those tasks. 

\subsection{Vision Transformer}
Encoder-only image-classification ViTs take as input a sequence of linear projections of equal-sized image patches with added position embedding and a special ``class token'' denoted [CLS]. Following the work of \citep{vilas2024analyzing} we use a version of the logit lens adapted to ViT to project the hidden state representations of each layer in the encoder onto the class embedding space using the output embedding matrix. Importantly this is done only for the [CLS] token for each image under the assumption that it best represents the model’s prediction, since during ViT’s pretraining the only token projected onto the class-embedding space is the [CLS] token from the last layer.

\begin{table}[tb!]
\caption{Accuracy of task number logistic regression classifier showing that in all modalities the layer embeddings contain information about the task number.}
\label{tab:probing_result}
\begin{center}
\resizebox{\columnwidth}{!}{%
\begin{tabular}{lrrr}
\toprule
\multicolumn{1}{c}{\bf Model}  &\multicolumn{1}{c}{\bf Layer Embeddings} &\multicolumn{1}{c}{\bf Random Embeddings}&\multicolumn{1}{c}{\bf Chance Level}
\\ \midrule
GPT2-XL (pretrained)  &$ \textbf{91.4} \pm 0.3 $ & $20.6 \pm 0.5$ & 20.0\\
GPT2-XL (random initialization) & $\textbf{86.1} \pm 0.7$ & $32.7 \pm 0.1$ & 33.3 \\
ViT-L/16 (pretrained)   & $\textbf{63.8} \pm 0.1 $& $21.0 \pm 0.5 $& 20.0  \\
Whisper-large (pretrained)  & $\textbf{52.7} \pm 0.1$ & $24.5 \pm 0.4$  & 25.0
\\\bottomrule
\end{tabular}
}
\end{center}
\end{table}

For our experiments we use the ViT-L/16 variant pretrained on ImageNet-21k and fine-tuned on ImageNet 2012, which has 1K classes and 24 layers, and run inference on 5K randomly sampled images from the CIFAR-10 \citep{krizhevsky2009learning} dataset. Figure \ref{fig:monton_vit} demonstrates the high correspondence between saturation layer and token rank, and the stricter Kendall's  {$\tau$} coefficient indicates a moderate agreement between the saturation layers order and the sequence $(1,2,..,k) $ which is statistically significant with $p<0.001$ (see Appendix \ref{sec:apx_kendal_tau}), supporting our claim that in this domain as well as in text the saturation layers are highly ordered. Furthermore, Table \ref{tab:probing_result} shows that the task index can be predicted from the hidden layer activations with high accuracy well above chance and control setting.

\subsection{Speech Transformer}
Whisper is an encoder-decoder Transformer model trained on many different speech processing tasks, including ASR. Although recently there have been attempts to increase efficiency in ASR, such as \citet{malard2023big}, the concept of early exit has yet to be explored in this setting, and to the best of our knowledge there has not been work done concerning saturation events in speech models. We adapt the logit lens and apply it \textit{only} to the decoder stack of Whisper-large, which has 32 layers, under the assumption that representations in the encoder stack are inherently different and projecting them onto the token vocabulary space would not be meaningful. For our dataset we randomly sample 5K audios from LibriSpeech \citep{panayotov2015librispeech}.

In addition to reproducing the classical top-1 saturation event established in language and vision models in previous work, we also show in Figure \ref{fig:monton_whisper} evidence for the tendency of the top-k tokens to reach saturation in order in this model as well, albeit only up to the third token. Moreover, a permutation test performed on the stricter Kendall's  {$\tau$} coefficient demonstrates that the agreement between the token ranking and the order of saturation layers is statistically significant with $p < 0.001$ (see Appendix \ref{sec:apx_kendal_tau}). We suspect that the order deteriorates in later tokens due to the fact that each layer in the decoder is conditioned on the last layer of the encoder which may interfere with the task transition mechanism by ``blurring the lines" between the tasks. Even so, Table \ref{tab:probing_result} shows that the task index can be predicted from Whisper's decoder layers' embeddings for tasks 1 through 4 with accuracy much higher than chance or that achieved in the control setting. 

\section{Analysis}
We have shown that top-k tokens tend to reach saturation in order of their ranking, as well as the plausibility of the underlying task transition mechanism over multiple modalities and variants of Transformers: decoder only, encoder only and full Transformer; in section \ref{untrained_transf} we argue that this phenomenon is inherent to the architecture itself and in section \ref{interv_proc} we delve deeper into the way the model transitions between tasks, demonstrating that we can cause the model to switch to the next task using an intervention procedure.

\subsection{Untrained Transformers Also Determine in Order}
\label{untrained_transf}
We repeat our experiments on an untrained  GPT2-XL with randomly initialized weights on the same amount of randomly sampled tokens from CNN/DM dataset as with the pretrained model.  Surprisingly, Figure \ref{fig:monton_random} shows that the top-k tokens tend to reach saturation in order up to the 4th token, and although the stricter Kendall's $\tau$ coefficient is lower than in the pretrained GPT2-XL model (see Appendix \ref{sec:apx_kendal_tau}), it is still statistically significant. 

\begin{figure}[t!]
\begin{center}
    \includegraphics[width=0.9\linewidth]{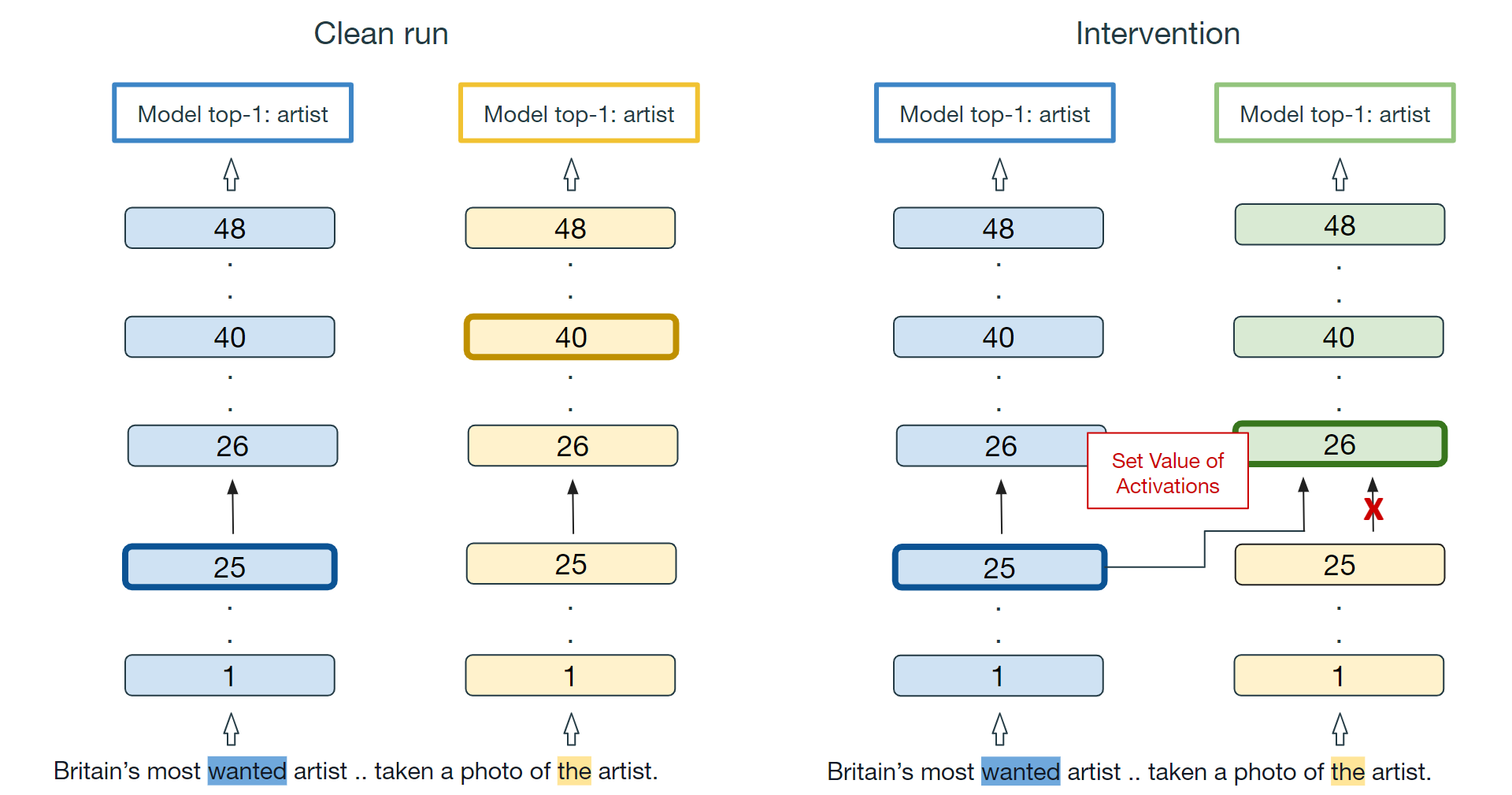} 
\end{center}
\caption{\emph{Left}: Forward pass of two input tokens (``wanted" and ``the") in the same context for which the model's final top-1 prediction is the same (``artist"), but the 1st saturation layers are different (25 and 40 respectively). \emph{Right}: By injecting the output from the top-1 saturation layer of ``the" as input to the subsequent layer of ``artist",  we trigger a saturation at the injected layer (26) in the post-intervention run, without altering the top-1 prediction. Saturation layers are marked in bold. The use of activations from adjacent layers is not depicted for the sake of clarity.}
\label{fig:interv_proc_vis}
\end{figure}

In addition, Table \ref{tab:probing_result} shows that the task transition classifier's accuracy is more than $2.5x$ times higher than chance or that of the control setting. The ability of the classifier to extract the task index from the hidden layers' representations in this setting is especially remarkable, demonstrating that despite the randomness of the weights as well of the identities of the predicted tokens, there is still highly ordered information encoded in the model originating only from the constraints of the architecture.
 
\subsection{Intervening in Layer Activations Causes Task Switch}
\label{interv_proc}
Using the probing analysis, we demonstrated that the tasks, as we defined them, are distinct enough to be separated by a simple classifier, that saturation layers mark the boundaries between them, and that the task index is encoded in the hidden layer embeddings. We argue that in addition, each saturation layer encodes the signal to transition to the next task, and all subsequent layers contain the information that the previous task has been completed and that the relevant token is fixed. This can be thought of as switch being flipped ``on" for each token that reaches saturation, and remaining "on" from the saturation layer onwards. 

\begin{figure}[tb!]
\begin{center}
\includegraphics[width=\linewidth]{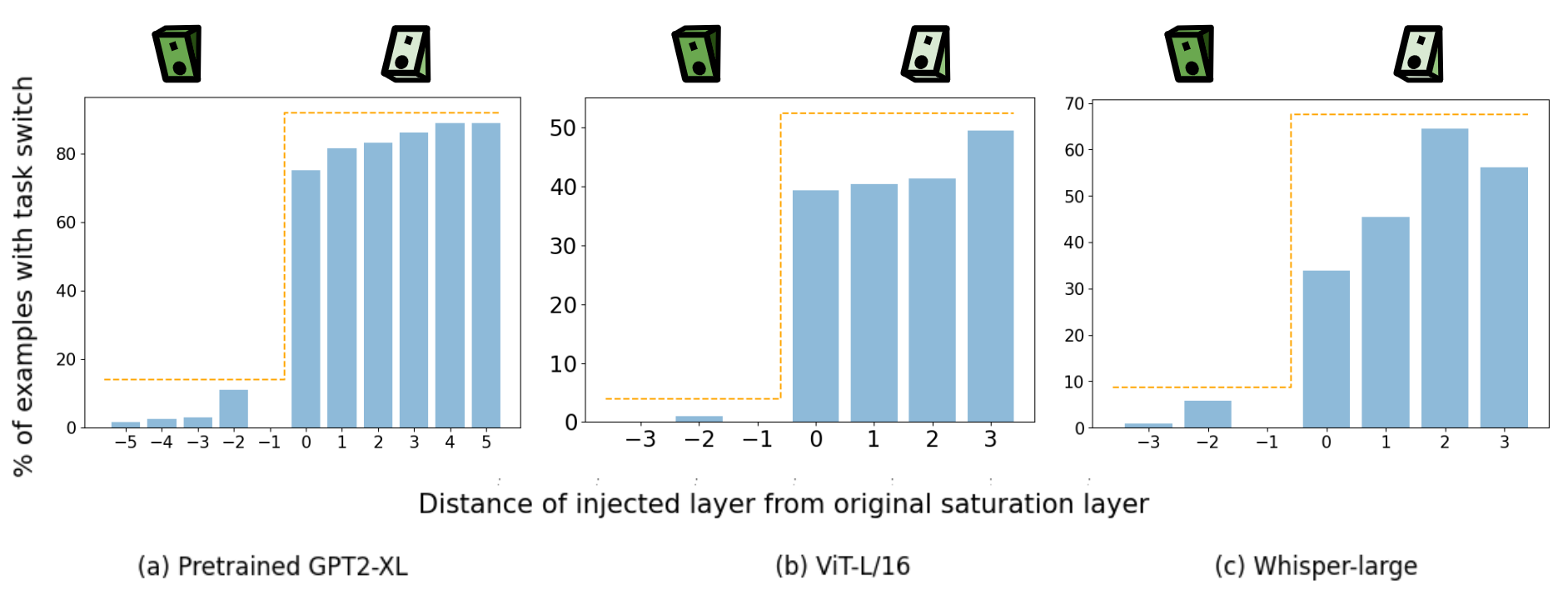} 
\end{center}
\caption{\emph{Flipping the Top-1 Switch.} The percentage of examples where the top-1 saturation occurred at the injected layer after the intervention, shown as a function of the layer from which the injected activations were taken, relative to the original saturation layer (e.g., $-2$ means activations were taken from two layers before the original saturation layer).}
\label{fig:inject-result}
\end{figure}

To causally validate this claim, taking inspiration from \citet{stolfo2023mechanistic}, we perform an intervention (visualized in Figure \ref{fig:interv_proc_vis}) in which we ``inject" the output from the 1st saturation layer of sample $s_1$ as input into the subsequent layer in the run on sample $s_2$ and check how this affects the 1st saturation layer of $s_2$. If these activations contain the signal to switch to the next task, we expect this intervention to cause the model to switch to task 2 at the injected layer in the post-intervention run, i.e. in the new post-intervention run the 1st saturation layer should be the one on which the intervention is performed which is $l^1(s_1)+1$. To minimize the effect of confounding factors, we choose pairs of samples $s_1$ and $s_2$ that share context and where the original top-1 prediction of the model is the same, but there is a big difference in their 1st saturation layers s.t. $l^1(s_1) <l^2(s_2)$. In the example depicted in Figure \ref{fig:interv_proc_vis} $s_1 = ``wanted"$ and $s_2 =``the"$, and for both the model's top-1 next-word prediction is ``artist", but $l^1(s_1)=25$ while $l^1(s_2)=40$. Injecting the output of layer $l^1(s_1)$ into the subsequent layer (26) in the run for $s_2$ should cause the model to switch to task 2, resulting in layer 26 being the new 1st saturation layer post intervention. 

Moreover, we would expect the same thing to happen when injecting activations from a layer $l$ \textit{after} the 1st saturation layer, i.e. $l>l^1(s_1)$, since they should contain the information that the top-1 token is fixed. On the other hand, activations from a layer $l'$ \textit{before} the 1st saturation layer, i.e. $l'<l^1(s_1)$ should not result in saturation at the injected layer as the switch is still ``off" in our analogy, indicating to the model that it still working on task 1. To test this, we repeat the same steps with activations from 5 layers before and after the 1st saturation layer $[l^1(s_1)-5, l^1(s_1)+5]$ each time injecting them as input into the subsequent layer.

Figure \ref{fig:inject-result}(a) shows the results of this procedure performed using pretrained GPT2-XL on 200 token pairs taken from 5 randomly sampled texts from the CNN/DM dataset,  Figure~\ref{fig:inject-result}(b) shows similar results reproduced using ViT-L/16 on 200 pairs of images from CIFAR-10 dataset, and Figure \ref{fig:inject-result}(c) shows the results of the intervention on Whisper-large on 200 token pairs from 100 randomly sampled audios from LibriSpeech \footnote{See Appendix \ref{sec:interv_proc_det} for a formal description of the procedure as well as details on how we adapted it for vision and speech modalities}. There is a stark difference in the effect the injected activations have on the 1st saturation layer post-intervention when the activations are taken from the 1st saturation layer in the original run or one of the following layers, compared to the layers before it. When the injected activations are taken from an earlier layer, the new top-1 saturation almost never occurs at the injected layer, whereas when the injected activations are taken from the saturation layer or a later layer the top-1 saturation occurs at the injected layer in a high percentage of cases. This drastic change resembles a step function, and is in line with our description of a switch being flipped "on" at the 1st saturation layer and remaining turned on in all subsequent layers, indicating to the model to switch to the next task and keep the top-1 constant.

 \section{Practical Applications}
 In this section, we show that our findings can be leveraged for computation efficiency and better performance in LLMs.

 \subsection{New Early-Exit Strategy}
We propose a new token-level dynamic inference method based on the task-transition classifier described in Section \ref{sec:transition-exp}, where the early exit layer for each token is defined as the earliest layer which is predicted to belong to task 2 by the classifier. The idea being that once the model has transitioned into the second task, it has finished with the first task of determining the top-1 token.

To demonstrate the viability and advantages of this method, we compare it to two other local confidence measures introduced by \citet{schuster2022confident}: softmax response (the difference between the top two values of the logits after softmax) and hidden-state saturation (cosine similarity between two consecutive layer embeddings), both recently found to be competitive with other early exiting methods~\citep{zhou2024survey}. Since dynamic decoding is not the focus of this paper, we calculate the metrics for each measure while propagating states from the layers after the “early exit” as in regular inference.\footnote{This is an informative comparison between the measures, as the effect of a state copying mechanism for skipped layers on model's performance is negligible \citet{schuster2022confident}. } 

Table \ref{tab:early_exit} shows our results on a pre-trained GPT2-XL model and 100 randomly sampled texts from CNN/DM dataset. We evaluate the model on next-word prediction, and compute the speedup ratio for each method as the number of layers it uses for each token divided by the total number of layers in the model, and average across all tokens. For the two other local confidence measures we calculate these metrics at various thresholds (see details in Appendix \ref{sec:early-exit_detail}), while in our metric the class is selected based on the highest predicted score among all classes. Our method outperforms the other two when considering the trade-off between next-word prediction accuracy and speedup ratio, and requires no training besides that of a simple logistic regression classifier on a relatively small amount of data.  

\begin{table}[tb!]
\caption{Highest accuracy and corresponding speedup-ratio achieved by each early-exit strategy.}
\label{tab:early_exit}
\begin{center}
\small
\begin{tabular}{lccc}
\toprule
\multicolumn{1}{c}{ }  &\multicolumn{1}{c}{\bf Softmax Response} &\multicolumn{1}{c}{\bf State Saturation}&\multicolumn{1}{c}{\bf Ours}
\\ \midrule
\textbf{Accuracy}  &$ 35.9 \pm 0.6 $ & $37.5\pm 0.7$ & $\textbf{40.0} \pm 0.7$\\
\textbf{Speedup Ratio} & $1.126 \pm 0.004 $ & $1.003 \pm 0.001$ & $\textbf{1.185} \pm 0.009$ 
\\\bottomrule
\end{tabular}
\end{center}
\end{table}

\subsection{Improved Language Modeling}
Popular decoding methods in language generation such as top-k  \citep{fan2018hierarchical} or top-p \citep{holtzman2020curious} sample the next token according to the shifted distribution induced by probabilities of the top ranking tokens. Based on our task-transition mechanism and the assumption that the tasks represent relevant computation, we argue that top ranking tokens that are determined in the last layer represent less meaningful predictions, since the model only had enough layers for the first task in these instances.

To test this hypothesis we compare the accuracy of the second ranked token in the next word prediction task between two conditions: (1) the second token’s saturation layer is at least 7 layers before the output, to increase the chances that this is a ``true" saturation as the model had enough layers to change its (recurring error) prediction and it is not due to noise; (2) the second token does not saturate, and is determined only in the last layer. In both cases we only look at examples where the top-1 token is not the correct prediction. The number of layers in the first setting is a hyper-parameter, and future work should investigate its affects on the second ranked token accuracy.

Using 100 randomly sampled texts from CNN/DM dataset and pretrained GPT2-XL predictions, we find that in the first condition the accuracy is 31.99\%, and in the second condition it is only 17.14\%. A Two Proportion Z-Test indicates a statistically significant difference between the groups ($p<0.001$). This supports our claim that top-k tokens that reach true saturation are more plausible than those that are determined only in the last layer, which has potential implications for generation decoding strategies which consider tokens beyond the top-1.

 \section{Related Work}
 There are multiple ways of thinking about the role of intermediate layers in Deep Neural Networks (DNNs) in general, and Transformers in particular. The iterative inference hypothesis interprets each layer as an iteration from an iterative and convergent process \citep{simoulin2021how}, suggesting that each layer incrementally refines the hidden representation by gradually shaping the next token prediction \citep {geva2022transformer, belrose2023eliciting, rushing2024explorations}. We argue that our findings challenge this view, given the discrete nature of the tasks in the proposed task-transition mechanism and the sharp transitions between them.

Pruning is another approach, focused on mitigating the redundancy inherent to large machine learning models by removing unnecessary parameters. Recent work has applied structured pruning methods to Transformer based LLMs, dropping whole modules, from self-attention layers \citep{artzy2024attend, he2024matters} to full Transformer blocks \citep{sun2024transformer, men2024shortgpt}. These studies often focus on the middle layers of the model, and claim to reduce memory and computation costs without degrading performance on downstream tasks. It’s important to note that these works evaluate the accuracy before and after pruning based only on the top-1 prediction of the model, even though stochastic generation methods such as top-p \citep{holtzman2019curious} and top-k \citep{fan2018hierarchical} are preferable to deterministic decoding in certain settings such as open-ended tasks as they produce more coherent and varied text \citep{shi2024thorough}. In light of this and of our results regarding the sequential saturation of top ranking tokens, we suggest that future work takes this into account, since what may seem as redundancy is actually necessary computation that is not reflected in the measured metric. 

The logit lens has also been used to study intermediate layers in a wide variety of interpretability papers \citep{yang2024large, wendler2024llamas, halawi2023overthinking}. Despite this, \cite {belrose2023eliciting} claim that it can produce implausible results due to the difference in representations between layers. To address this issue they introduce the ``tuned lens", in which an affine transformation is learned for each layer in the model with a distillation loss, so that its image under the unembedding matrix matches the final layer logits as closely as possible. Although this method may be better at approximating final top-1 prediction from intermediate layers, our work highlights why this might actually be a disadvantage when attempting to gain insights into the computational process of the Transformer, as it could obscure the changing dynamics of the lower ranked tokens. 

 \section{Conclusion and Future Work}
 This paper systematically investigates the unexplored question of what computation is performed by the Transformer layers following a top-1 saturation event. We find that the top-k tokens (for $k>1$)  go through similar saturation events in the order of their ranking. We argue that this phenomenon is inherent to the Transformer architecture, replicating our results on an untrained model and demonstrating its robustness over multiple modalities: text, vision, and (to a lesser extent) speech. We then provide evidence in support of a task transition as underlying mechanism for this ordered saturation, showing that we can predict task index from the hidden layers' embeddings, as well as cause the model to switch from the first task to the second via an intervention procedure. Our findings also hold promise in improving inference efficiency and next word prediction accuracy as suggested by the preliminary results in the Practical Applications section. 


\textbf{Limitations and Future Work}
Although our analysis sheds light on the high-level task transition mechanism behind the ordered saturation of top-k tokens, there is still a need for more work to determine which components in the Transformer architecture give rise to it, via ablation studies for example, as well as more concrete explanation for how the model keeps the “chosen” tokens constant after their saturation events across remaining layers. In addition, we did not consider whether the model encountered the data used in our experiments during training as a relevant factor. Finally, as we only explored Transformer architectures, it is necessary to check whether other types of DNNs also determine their top-k tokens in order. Recurrent Neural Networks (RNNs) might be of particular interest due to their mathematical equivalence to decoder only Transformers \citep {oren2024transformers}, and based on previous work successfully applying the logit lens to them to extract meaningful predictions from intermediate layers \citep{paulo2024does}.

 \section{Acknowledgements}
 We would like to thank Nir Rosenfeld, Omri Abend and Mariano Schain for the invaluable insights, as well as Noam Dahan and Timna Wharton Kleinman for their helpful feedback.

\bibliographystyle{iclr2025_conference}
\bibliography{references}

\begin{thebibliography}{34}
\providecommand{\natexlab}[1]{#1}
\providecommand{\url}[1]{\texttt{#1}}
\expandafter\ifx\csname urlstyle\endcsname\relax
  \providecommand{\doi}[1]{doi: #1}\else
  \providecommand{\doi}{doi: \begingroup \urlstyle{rm}\Url}\fi

\bibitem[Artzy \& Schwartz(2024)Artzy and Schwartz]{artzy2024attend}
Amit~Ben Artzy and Roy Schwartz.
\newblock Attend first, consolidate later: On the importance of attention in different llm layers.
\newblock \emph{arXiv preprint arXiv:2409.03621}, 2024.

\bibitem[Belrose et~al.(2023)Belrose, Furman, Smith, Halawi, Ostrovsky, McKinney, Biderman, and Steinhardt]{belrose2023eliciting}
Nora Belrose, Zach Furman, Logan Smith, Danny Halawi, Igor Ostrovsky, Lev McKinney, Stella Biderman, and Jacob Steinhardt.
\newblock Eliciting latent predictions from transformers with the tuned lens.
\newblock \emph{arXiv preprint arXiv:2303.08112}, 2023.

\bibitem[Dosovitskiy et~al.(2020)Dosovitskiy, Beyer, Kolesnikov, Weissenborn, Zhai, Unterthiner, and Houlsby]{dosovitskiy2020image}
Alexey Dosovitskiy, Lucas Beyer, Alexander Kolesnikov, Dirk Weissenborn, Xiaohua Zhai, Thomas Unterthiner, and Neil Houlsby.
\newblock An image is worth 16x16 words: Transformers for image recognition at scale.
\newblock \emph{arXiv preprint arXiv:2010.11929}, 2020.

\bibitem[Fan et~al.(2018)Fan, Lewis, and Dauphin]{fan2018hierarchical}
Angela Fan, Mike Lewis, and Yann Dauphin.
\newblock Hierarchical neural story generation.
\newblock \emph{arXiv preprint arXiv:1805.04833}, 2018.

\bibitem[Frydenlund et~al.(2022)Frydenlund, Singh, and Rudzicz]{frydenlund2022language}
Adrian Frydenlund, Gagandeep Singh, and Frank Rudzicz.
\newblock Language modelling via learning to rank.
\newblock In \emph{Proceedings of the AAAI conference on artificial intelligence}, volume~36, pp.\  10636--10644. AAAI, 2022.

\bibitem[Geva et~al.(2022)Geva, Caciularu, Wang, and Goldberg]{geva2022transformer}
Mor Geva, Avi Caciularu, Kevin~R Wang, and Yoav Goldberg.
\newblock Transformer feed-forward layers build predictions by promoting concepts in the vocabulary space.
\newblock \emph{arXiv preprint arXiv:2203.14680}, 2022.

\bibitem[Halawi et~al.(2023)Halawi, Denain, and Steinhardt]{halawi2023overthinking}
Danny Halawi, Jean-Stanislas Denain, and Jacob Steinhardt.
\newblock Overthinking the truth: Understanding how language models process false demonstrations.
\newblock \emph{arXiv preprint arXiv:2307.09476}, 2023.

\bibitem[He et~al.(2024)He, Sun, Shen, and Li]{he2024matters}
Shwai He, Guoheng Sun, Zheyu Shen, and Ang Li.
\newblock What matters in transformers? not all attention is needed.
\newblock \emph{arXiv preprint arXiv:2406.15786}, 2024.

\bibitem[Hermann et~al.(2015)Hermann, Kocisky, Grefenstette, Espeholt, Kay, Suleyman, and Blunsom]{hermann2015teaching}
Karl~Moritz Hermann, Tom Kocisky, Edward Grefenstette, Lasse Espeholt, Will Kay, Mustafa Suleyman, and Phil Blunsom.
\newblock Teaching machines to read and comprehend.
\newblock \emph{Advances in neural information processing systems}, 28, 2015.

\bibitem[Holtzman et~al.(2019)Holtzman, Buys, Du, Forbes, and Choi]{holtzman2019curious}
Ari Holtzman, Jan Buys, Li~Du, Maxwell Forbes, and Yejin Choi.
\newblock The curious case of neural text degeneration.
\newblock \emph{arXiv preprint arXiv:1904.09751}, 2019.

\bibitem[Holtzman et~al.(2020)Holtzman, Buys, Du, Forbes, and Choi]{holtzman2020curious}
Ari Holtzman, Jan Buys, Li~Du, Maxwell Forbes, and Yejin Choi.
\newblock The curious case of neural text degeneration.
\newblock In \emph{International Conference on Learning Representations (ICLR)}, 2020.

\bibitem[Krizhevsky et~al.(2009)Krizhevsky, Hinton, et~al.]{krizhevsky2009learning}
Alex Krizhevsky, Geoffrey Hinton, et~al.
\newblock Learning multiple layers of features from tiny images.
\newblock 2009.

\bibitem[Malard et~al.(2023)Malard, Zaiem, and Algayres]{malard2023big}
Hugo Malard, Soumaya Zaiem, and Robin Algayres.
\newblock Big model only for hard audios: Sample dependent whisper model selection for efficient inferences.
\newblock \emph{arXiv preprint arXiv:2309.12712}, 2023.

\bibitem[Men et~al.(2024)Men, Xu, Zhang, Wang, Lin, Lu, Han, and Chen]{men2024shortgpt}
Xin Men, Mingyu Xu, Qingyu Zhang, Bingning Wang, Hongyu Lin, Yaojie Lu, Xianpei Han, and Weipeng Chen.
\newblock Shortgpt: Layers in large language models are more redundant than you expect.
\newblock \emph{arXiv preprint arXiv:2403.03853}, 2024.

\bibitem[Nostalgebraist(2020)]{nostalgebraist2020gpt}
Nostalgebraist.
\newblock Interpreting gpt: the logit lens, 2020.

\bibitem[OpenAI et~al.(2024)OpenAI, Achiam, Adler, Agarwal, Ahmad, Akkaya, Aleman, Almeida, Altenschmidt, Altman, Anadkat, Avila, Babuschkin, Balaji, Balcom, Baltescu, Bao, Bavarian, Belgum, Bello, Berdine, Bernadett-Shapiro, Berner, Bogdonoff, Boiko, Boyd, Brakman, Brockman, Brooks, Brundage, Button, Cai, Campbell, Cann, Carey, Carlson, Carmichael, Chan, Chang, Chantzis, Chen, Chen, Chen, Chen, Chen, Chess, Cho, Chu, Chung, Cummings, Currier, Dai, Decareaux, Degry, Deutsch, Deville, Dhar, Dohan, Dowling, Dunning, Ecoffet, Eleti, Eloundou, Farhi, Fedus, Felix, Fishman, Forte, Fulford, Gao, Georges, Gibson, Goel, Gogineni, Goh, Gontijo-Lopes, Gordon, Grafstein, Gray, Greene, Gross, Gu, Guo, Hallacy, Han, Harris, He, Heaton, Heidecke, Hesse, Hickey, Hickey, Hoeschele, Houghton, Hsu, Hu, Hu, Huizinga, Jain, Jain, Jang, Jiang, Jiang, Jin, Jin, Jomoto, Jonn, Jun, Kaftan, Łukasz Kaiser, Kamali, Kanitscheider, Keskar, Khan, Kilpatrick, Kim, Kim, Kim, Kirchner, Kiros, Knight, Kokotajlo, Łukasz Kondraciuk, Kondrich,
  Konstantinidis, Kosic, Krueger, Kuo, Lampe, Lan, Lee, Leike, Leung, Levy, Li, Lim, Lin, Lin, Litwin, Lopez, Lowe, Lue, Makanju, Malfacini, Manning, Markov, Markovski, Martin, Mayer, Mayne, McGrew, McKinney, McLeavey, McMillan, McNeil, Medina, Mehta, Menick, Metz, Mishchenko, Mishkin, Monaco, Morikawa, Mossing, Mu, Murati, Murk, Mély, Nair, Nakano, Nayak, Neelakantan, Ngo, Noh, Ouyang, O'Keefe, Pachocki, Paino, Palermo, Pantuliano, Parascandolo, Parish, Parparita, Passos, Pavlov, Peng, Perelman, de~Avila Belbute~Peres, Petrov, de~Oliveira~Pinto, Michael, Pokorny, Pokrass, Pong, Powell, Power, Power, Proehl, Puri, Radford, Rae, Ramesh, Raymond, Real, Rimbach, Ross, Rotsted, Roussez, Ryder, Saltarelli, Sanders, Santurkar, Sastry, Schmidt, Schnurr, Schulman, Selsam, Sheppard, Sherbakov, Shieh, Shoker, Shyam, Sidor, Sigler, Simens, Sitkin, Slama, Sohl, Sokolowsky, Song, Staudacher, Such, Summers, Sutskever, Tang, Tezak, Thompson, Tillet, Tootoonchian, Tseng, Tuggle, Turley, Tworek, Uribe, Vallone, Vijayvergiya,
  Voss, Wainwright, Wang, Wang, Wang, Ward, Wei, Weinmann, Welihinda, Welinder, Weng, Weng, Wiethoff, Willner, Winter, Wolrich, Wong, Workman, Wu, Wu, Wu, Xiao, Xu, Yoo, Yu, Yuan, Zaremba, Zellers, Zhang, Zhang, Zhao, Zheng, Zhuang, Zhuk, and Zoph]{openai2024gpt4technicalreport}
OpenAI, Josh Achiam, Steven Adler, Sandhini Agarwal, Lama Ahmad, Ilge Akkaya, Florencia~Leoni Aleman, Diogo Almeida, Janko Altenschmidt, Sam Altman, Shyamal Anadkat, Red Avila, Igor Babuschkin, Suchir Balaji, Valerie Balcom, Paul Baltescu, Haiming Bao, Mohammad Bavarian, Jeff Belgum, Irwan Bello, Jake Berdine, Gabriel Bernadett-Shapiro, Christopher Berner, Lenny Bogdonoff, Oleg Boiko, Madelaine Boyd, Anna-Luisa Brakman, Greg Brockman, Tim Brooks, Miles Brundage, Kevin Button, Trevor Cai, Rosie Campbell, Andrew Cann, Brittany Carey, Chelsea Carlson, Rory Carmichael, Brooke Chan, Che Chang, Fotis Chantzis, Derek Chen, Sully Chen, Ruby Chen, Jason Chen, Mark Chen, Ben Chess, Chester Cho, Casey Chu, Hyung~Won Chung, Dave Cummings, Jeremiah Currier, Yunxing Dai, Cory Decareaux, Thomas Degry, Noah Deutsch, Damien Deville, Arka Dhar, David Dohan, Steve Dowling, Sheila Dunning, Adrien Ecoffet, Atty Eleti, Tyna Eloundou, David Farhi, Liam Fedus, Niko Felix, Simón~Posada Fishman, Juston Forte, Isabella Fulford, Leo
  Gao, Elie Georges, Christian Gibson, Vik Goel, Tarun Gogineni, Gabriel Goh, Rapha Gontijo-Lopes, Jonathan Gordon, Morgan Grafstein, Scott Gray, Ryan Greene, Joshua Gross, Shixiang~Shane Gu, Yufei Guo, Chris Hallacy, Jesse Han, Jeff Harris, Yuchen He, Mike Heaton, Johannes Heidecke, Chris Hesse, Alan Hickey, Wade Hickey, Peter Hoeschele, Brandon Houghton, Kenny Hsu, Shengli Hu, Xin Hu, Joost Huizinga, Shantanu Jain, Shawn Jain, Joanne Jang, Angela Jiang, Roger Jiang, Haozhun Jin, Denny Jin, Shino Jomoto, Billie Jonn, Heewoo Jun, Tomer Kaftan, Łukasz Kaiser, Ali Kamali, Ingmar Kanitscheider, Nitish~Shirish Keskar, Tabarak Khan, Logan Kilpatrick, Jong~Wook Kim, Christina Kim, Yongjik Kim, Jan~Hendrik Kirchner, Jamie Kiros, Matt Knight, Daniel Kokotajlo, Łukasz Kondraciuk, Andrew Kondrich, Aris Konstantinidis, Kyle Kosic, Gretchen Krueger, Vishal Kuo, Michael Lampe, Ikai Lan, Teddy Lee, Jan Leike, Jade Leung, Daniel Levy, Chak~Ming Li, Rachel Lim, Molly Lin, Stephanie Lin, Mateusz Litwin, Theresa Lopez, Ryan
  Lowe, Patricia Lue, Anna Makanju, Kim Malfacini, Sam Manning, Todor Markov, Yaniv Markovski, Bianca Martin, Katie Mayer, Andrew Mayne, Bob McGrew, Scott~Mayer McKinney, Christine McLeavey, Paul McMillan, Jake McNeil, David Medina, Aalok Mehta, Jacob Menick, Luke Metz, Andrey Mishchenko, Pamela Mishkin, Vinnie Monaco, Evan Morikawa, Daniel Mossing, Tong Mu, Mira Murati, Oleg Murk, David Mély, Ashvin Nair, Reiichiro Nakano, Rajeev Nayak, Arvind Neelakantan, Richard Ngo, Hyeonwoo Noh, Long Ouyang, Cullen O'Keefe, Jakub Pachocki, Alex Paino, Joe Palermo, Ashley Pantuliano, Giambattista Parascandolo, Joel Parish, Emy Parparita, Alex Passos, Mikhail Pavlov, Andrew Peng, Adam Perelman, Filipe de~Avila Belbute~Peres, Michael Petrov, Henrique~Ponde de~Oliveira~Pinto, Michael, Pokorny, Michelle Pokrass, Vitchyr~H. Pong, Tolly Powell, Alethea Power, Boris Power, Elizabeth Proehl, Raul Puri, Alec Radford, Jack Rae, Aditya Ramesh, Cameron Raymond, Francis Real, Kendra Rimbach, Carl Ross, Bob Rotsted, Henri Roussez,
  Nick Ryder, Mario Saltarelli, Ted Sanders, Shibani Santurkar, Girish Sastry, Heather Schmidt, David Schnurr, John Schulman, Daniel Selsam, Kyla Sheppard, Toki Sherbakov, Jessica Shieh, Sarah Shoker, Pranav Shyam, Szymon Sidor, Eric Sigler, Maddie Simens, Jordan Sitkin, Katarina Slama, Ian Sohl, Benjamin Sokolowsky, Yang Song, Natalie Staudacher, Felipe~Petroski Such, Natalie Summers, Ilya Sutskever, Jie Tang, Nikolas Tezak, Madeleine~B. Thompson, Phil Tillet, Amin Tootoonchian, Elizabeth Tseng, Preston Tuggle, Nick Turley, Jerry Tworek, Juan Felipe~Cerón Uribe, Andrea Vallone, Arun Vijayvergiya, Chelsea Voss, Carroll Wainwright, Justin~Jay Wang, Alvin Wang, Ben Wang, Jonathan Ward, Jason Wei, CJ~Weinmann, Akila Welihinda, Peter Welinder, Jiayi Weng, Lilian Weng, Matt Wiethoff, Dave Willner, Clemens Winter, Samuel Wolrich, Hannah Wong, Lauren Workman, Sherwin Wu, Jeff Wu, Michael Wu, Kai Xiao, Tao Xu, Sarah Yoo, Kevin Yu, Qiming Yuan, Wojciech Zaremba, Rowan Zellers, Chong Zhang, Marvin Zhang, Shengjia
  Zhao, Tianhao Zheng, Juntang Zhuang, William Zhuk, and Barret Zoph.
\newblock Gpt-4 technical report, 2024.
\newblock URL \url{https://arxiv.org/abs/2303.08774}.

\bibitem[Oren et~al.(2024)Oren, Hassid, Adi, and Schwartz]{oren2024transformers}
Matanel Oren, Michael Hassid, Yossi Adi, and Roy Schwartz.
\newblock Transformers are multi-state rnns.
\newblock \emph{arXiv preprint arXiv:2401.06104}, 2024.

\bibitem[Panayotov et~al.(2015)Panayotov, Chen, Povey, and Khudanpur]{panayotov2015librispeech}
Vassil Panayotov, Guoguo Chen, Daniel Povey, and Sanjeev Khudanpur.
\newblock Librispeech: an asr corpus based on public domain audio books.
\newblock In \emph{2015 IEEE international conference on acoustics, speech and signal processing (ICASSP)}, pp.\  5206--5210. IEEE, 2015.

\bibitem[Paulo et~al.(2024)Paulo, Marshall, and Belrose]{paulo2024does}
Gon{\c{c}}alo Paulo, Thomas Marshall, and Nora Belrose.
\newblock Does transformer interpretability transfer to rnns?
\newblock \emph{arXiv preprint arXiv:2404.05971}, 2024.

\bibitem[Radford et~al.(2019)Radford, Wu, Child, Luan, Amodei, and Sutskever]{radford2019language}
Alec Radford, Jeffrey Wu, Rewon Child, David Luan, Dario Amodei, and Ilya Sutskever.
\newblock Language models are unsupervised multitask learners.
\newblock \emph{OpenAI blog}, 1\penalty0 (8):\penalty0 9, 2019.

\bibitem[Radford et~al.(2023)Radford, Kim, Xu, Brockman, McLeavey, and Sutskever]{radford2023robust}
Alec Radford, Jong~Wook Kim, Tao Xu, Greg Brockman, Christine McLeavey, and Ilya Sutskever.
\newblock Robust speech recognition via large-scale weak supervision.
\newblock In \emph{International conference on machine learning}, pp.\  28492--28518. PMLR, 2023.

\bibitem[Rushing \& Nanda(2024)Rushing and Nanda]{rushing2024explorations}
Cody Rushing and Neel Nanda.
\newblock Explorations of self-repair in language models.
\newblock \emph{arXiv preprint arXiv:2402.15390}, 2024.

\bibitem[Schuster et~al.(2022)Schuster, Fisch, Gupta, Dehghani, Bahri, Tran, Tay, and Metzler]{schuster2022confident}
Tal Schuster, Adam Fisch, Jai Gupta, Mostafa Dehghani, Dara Bahri, Vu~Tran, Yi~Tay, and Donald Metzler.
\newblock Confident adaptive language modeling.
\newblock \emph{Advances in Neural Information Processing Systems}, 35:\penalty0 17456--17472, 2022.

\bibitem[Schwartz et~al.(2020)Schwartz, Stanovsky, Swayamdipta, Dodge, and Smith]{schwartz2020right}
Roy Schwartz, Gabriel Stanovsky, Swabha Swayamdipta, Jesse Dodge, and Noah~A Smith.
\newblock The right tool for the job: Matching model and instance complexities.
\newblock \emph{arXiv preprint arXiv:2004.07453}, 2020.

\bibitem[Shi et~al.(2024)Shi, Yang, Cai, Zhang, Wang, Yang, and Lam]{shi2024thorough}
Chufan Shi, Haoran Yang, Deng Cai, Zhisong Zhang, Yifan Wang, Yujiu Yang, and Wai Lam.
\newblock A thorough examination of decoding methods in the era of llms.
\newblock \emph{arXiv preprint arXiv:2402.06925}, 2024.

\bibitem[Simoulin \& Crabb{\'e}(2021)Simoulin and Crabb{\'e}]{simoulin2021how}
Alex Simoulin and Beno{\^i}t Crabb{\'e}.
\newblock How many layers and why? an analysis of the model depth in transformers.
\newblock In \emph{Proceedings of the 59th Annual Meeting of the Association for Computational Linguistics and the 11th International Joint Conference on Natural Language Processing: Student Research Workshop}, pp.\  221--228, 2021.

\bibitem[Stolfo et~al.(2023)Stolfo, Belinkov, and Sachan]{stolfo2023mechanistic}
Andrew Stolfo, Yonatan Belinkov, and Mrinmaya Sachan.
\newblock A mechanistic interpretation of arithmetic reasoning in language models using causal mediation analysis.
\newblock \emph{arXiv preprint arXiv:2305.15054}, 2023.

\bibitem[Sun et~al.(2024)Sun, Pickett, Nain, and Jones]{sun2024transformer}
Qi~Sun, Marc Pickett, Aakash~Kumar Nain, and Llion Jones.
\newblock Transformer layers as painters.
\newblock \emph{arXiv preprint arXiv:2407.09298}, 2024.

\bibitem[Vaswani et~al.(2017)]{vaswani2017attention}
Ashish Vaswani et~al.
\newblock Attention is all you need.
\newblock \emph{Advances in Neural Information Processing Systems}, 2017.

\bibitem[Vilas et~al.(2024)Vilas, Schauml{\"o}ffel, and Roig]{vilas2024analyzing}
Miquel~Gonzalez Vilas, Toni Schauml{\"o}ffel, and Gemma Roig.
\newblock Analyzing vision transformers for image classification in class embedding space.
\newblock \emph{Advances in Neural Information Processing Systems}, 36, 2024.

\bibitem[Wendler et~al.(2024)Wendler, Veselovsky, Monea, and West]{wendler2024llamas}
Chris Wendler, Veniamin Veselovsky, Giovanni Monea, and Robert West.
\newblock Do llamas work in english? on the latent language of multilingual transformers.
\newblock \emph{arXiv preprint arXiv:2402.10588}, 2024.

\bibitem[Yang et~al.(2024)Yang, Gribovskaya, Kassner, Geva, and Riedel]{yang2024large}
Sohee Yang, Elena Gribovskaya, Nora Kassner, Mor Geva, and Sebastian Riedel.
\newblock Do large language models latently perform multi-hop reasoning?
\newblock \emph{arXiv preprint arXiv:2402.16837}, 2024.

\bibitem[Zhang et~al.(2023)Zhang, Han, Qin, Wang, Bapna, Chen, Chen, Li, Axelrod, Wang, et~al.]{zhang2023google}
Yu~Zhang, Wei Han, James Qin, Yongqiang Wang, Ankur Bapna, Zhehuai Chen, Nanxin Chen, Bo~Li, Vera Axelrod, Gary Wang, et~al.
\newblock Google usm: Scaling automatic speech recognition beyond 100 languages.
\newblock \emph{arXiv preprint arXiv:2303.01037}, 2023.

\bibitem[Zhou et~al.(2024)Zhou, Ning, Hong, Fu, Xu, Li, Lou, Wang, Yuan, Li, et~al.]{zhou2024survey}
Zixuan Zhou, Xuefei Ning, Ke~Hong, Tianyu Fu, Jiaming Xu, Shiyao Li, Yuming Lou, Luning Wang, Zhihang Yuan, Xiuhong Li, et~al.
\newblock A survey on efficient inference for large language models.
\newblock \emph{arXiv preprint arXiv:2404.14294}, 2024.

\end{thebibliography}

\appendix
\section{Appendix}
\subsection{Stricter Kendall's tau}
\label{sec:apx_kendal_tau}
We define a version of Kendall's tau coefficient measuring the ordinal association between two tanking, where one-sided ties are considered discordant unlike the regular metric, where ties are typically either ignored or handled as neutral, meaning they neither count as concordant nor discordant.

Given two rankings $x=(x_1,x_2, ., x_n)$ and \ $y=(y_1,y_2, ., y_n)$, let \( \text{pair}(i, j) \) be a pair of indices where \( 1 \leq i < j \leq n \). 

We define the pair as \textit{concordant} if the rankings in both sequences agree, meaning:
\[
(x_i > x_j \, \text{and} \, y_i > y_j) \quad \text{or} \quad (x_i < x_j \, \text{and} \, y_i < y_j) \quad \text{or} \quad (x_i = x_j \, \text{and} \, y_i = y_j) 
\]

The pair is \textit{discordant} if:
\[
(x_i > x_j \, \text{and} \, y_i < y_j) \quad \text{or} \quad (x_i > x_j \, \text{and} \, y_i > y_j) \quad \text{or} \quad (x_i = x_j \, \text{and} \, y_i \neq y_j) \quad \text{or} \quad (x_i \neq x_j \, \text{and} \, y_i = y_j) 
\]

The coefficient \(\tau_{\text{strict}} \), is computed as:

\[
\tau_{\text{strict}} = \frac{C - D}{C + D},
\]

where \( C \) is the number of concordant pairs, and \( D \) is the number of discordant pairs (including ties), ranging in values between $[-1,1]$.

Table \ref{tab:k-tau_result} summarizes the results of this metric across the different models discussed in the paper, along with the p-values of the permutation test performed for the mean $\tau_{strict}$ for each.

\begin{table}[ht]
\caption{Stricter Kendall's tau coefficients and p-values for each model }
\label{tab:k-tau_result}
\begin{center}
\resizebox{\columnwidth}{!}{%
\begin{tabular}{lrrr}
\toprule
\multicolumn{1}{c}{\bf Model}  &\multicolumn{1}{c}{\bf $\tau_{strict}$ (avg $\pm$ ste)}
&\multicolumn{1}{c}{\bf $p_{value}$}
&\multicolumn{1}{c}{\bf $\tau_{strict} > 0$}
\\ \midrule
GPT2-XL (pre-trained)          &\( 0.187 \pm 0.004\) & \( < 0.001 \)  &\( 67.39\%\) \\
GPT2-XL (random initialization)              & \(0.082 \pm 0.009\) & \(< 0.001\)  &\( 49.48\%\)\\
ViT-L/16 (pre-trained)              & \(0.149 \pm 0.007\) & \(< 0.001\)  & \(58.94\%\)  \\
Whisper-large (pre-trained)  & \(0.210 \pm 0.009\) &\( < 0.001 \) & \( 63.78\%\)
\\\bottomrule
\end{tabular}
}
\end{center}
\end{table}

\subsection{Training Data for Task Transition Classifier}
Table \ref{tab:probing_data} shows from which layers we took embeddings to train the task-transition classifier for each model. 

\label{sec:train_data_probing}
\begin{table}[ht]
\small
\caption{Task transition probing per-class accuracy scores}
\begin{center}
\begin{tabular}{lrr}
\toprule
\multicolumn{1}{c}{\bf Model}  &\multicolumn{1}{c}{\bf Layers}
&\multicolumn{1}{c}{\bf Dataset size}
\\ \midrule
GPT2-XL (pre-trained)          &\(23-40\) &\( 6K\) \\
GPT2-XL (random initialization)              & \(31-41\) & \(2K\)\\
ViT-L/16 (pre-trained)              &\(16-21\) &\(2K\) \\
Whisper-large (pre-trained)              &\(29-32\) &\(4K\)
\\\bottomrule
\end{tabular}
\label{tab:probing_data}
\end{center}
\end{table}

\subsection{Per Class Metrics for Task Transition Classifier}
\label{sec:per_class_probing_metrics}

\begin{table}[ht]
\caption{Task transition probing per-class accuracy scores}
\begin{center}
\resizebox{\columnwidth}{!}{%
\begin{tabular}{lrrrrr}
\toprule
\multicolumn{1}{c}{\bf Model}  &\multicolumn{1}{c}{\bf Task 1}
&\multicolumn{1}{c}{\bf Task 2}
&\multicolumn{1}{c}{\bf Task 3}
&\multicolumn{1}{c}{\bf Task 4}
&\multicolumn{1}{c}{\bf Task 5}
\\ \midrule
GPT2-XL (pre-trained)          &\(0.862\) &\( 0.92\) &\(0.926\) &\(0.921\) &\(0.941\) \\
GPT2-XL (random initialization)              & \(0.782\) & \(0.83\)  &\( 0.972\) & \(-\) & \(-\)\\
ViT-L/16 (pre-trained)              &\( 0.646\) &\(0.618\) &\(0.603\) &\(0.644\) &\(0.678\)\\
Whisper-large (pre-trained)              &\( 0.520\) &\(0.528\) &\(0.484\) &\(0.576\) &\(-\)
\\\bottomrule
\end{tabular}
}
\end{center}
\end{table}

\begin{table}[ht]
\caption{Task transition probing per-class ROC-AUC scores}
\begin{center}
\resizebox{\columnwidth}{!}{%
\begin{tabular}{lrrrrr}
\toprule
\multicolumn{1}{c}{\bf Model}  &\multicolumn{1}{c}{\bf Task 1}
&\multicolumn{1}{c}{\bf Task 2}
&\multicolumn{1}{c}{\bf Task 3}
&\multicolumn{1}{c}{\bf Task 4}
&\multicolumn{1}{c}{\bf Task 5}
\\ \midrule
GPT2-XL (pre-trained)       & \(0.966\) & \(0.987\) & \(0.974\)& \(0.974\)& \(0.985\)\\
GPT2-XL (random initialization)              & \(0.929\) & \(0.951\)  &\( 0.995\) & \(-\) & \(-\)\\
ViT-L/16 (pre-trained)              &\( 0.855\) &\(0.821\) &\(0.809\) &\(0.829\) &\(0.866\)\\
Whisper-large (pre-trained)              &\( 0.777\) &\(0.767\) &\(0.695\) &\(0.765\) &\(-\)
\\\bottomrule
\end{tabular}
}
\end{center}
\end{table}

\subsection{Intervention Procedure Additional Details}
\label{sec:interv_proc_det}
Formally, this procedure consists of the following steps:

\begin{enumerate}
    \item Given an input sequence \( x=<x_1,...,x_t>\) we first pass it through the model  as in regular inference while storing the activation values at all hidden layers, i.e  \(h_i^l\) for all \(1\leq i \leq t\), \(1\leq l\leq N\).
    \item We calculate the saturation layer \(l_i^1\) of the 1st token for each token \(w_i\) in the text.
    \item We sample pairs of token indexes \(i,j\)  in the text that the satisfy the following conditions:
    \begin{enumerate}
        \item The distance between \(i\) and \(j\) is no more than 40 tokens, i.e. \(|i-j|\leq 40\).  \newline
        This is a precaution to minimize the effect of the difference in context on the model's predictions after intervention.
        \item The model’s top-1 prediction (in the final layer) for both indexes is the same token y, meaning \(y=argmax(softmax(Eh_i^N))=argmax(softmax(Eh_j^N)) \). \newline
        The goal here is to avoid a conflict in the top-1 predictions which could be a confounding factor. 
        \item There is a difference of at least 10 layers between the 1st token saturation layers of i and j, such that \(|l_i^1-l_j^1|\leq 10\), to ensure that the change in saturation layer after intervention is significant.
    \end{enumerate}

    For convenience’s sake we will assume in the remainder of the procedure description that \(l_i^1<l_j^1\), i.e that the saturation layer of the first index in the pair is smaller then that of the second index (even though both cases are allowed by our conditions).
    
    \item We perform 11 additional forward passes, each time injecting the output from layer  $l'$ in range \([l_i^1 -5, l_i^1 +5]\) at position $i$ as input into layer $l'+1$ at position j . The goal here is to to quantify the difference in effect between layers preceding the saturation event and those after it.
    \item We measure the causal effect of the intervention by calculating the percent of examples where the saturation layer of the 1st token after intervention \(\tilde l_j^1\) is the layer on which we intervened, i.e.  $\tilde l_j^1=l+1$.
    \end{enumerate}

For example, in the setting depicted in \ref{fig:interv_proc_vis} we would take the indexes of the marked tokens "wanted" and "the" as our pair, where the original top-1 prediction in both is "artist". The top-1 saturation layer in the clean run for "wanted" is layer 25, so we would inject activations from layers 20 to 30 one at a time as inputs into the corresponding subsequent layers in the run of token "the" (i.e. layers 21 to 31), and check for each one if the injected layer became the new top-1 saturation layer after the intervention.

\subsubsection{Intervention Procedure on ViT} 
To adapt the intervention procedure described in section \ref{interv_proc} to ViT-L/16 and the image classification setting we made the following modifications:
\begin{enumerate}
    \item Since each image is processed independently by the model there is no need for two images to share a context, so the only requirements for two images to be chosen as a relevant pair were: a distance of at least 5 layers (instead of 10, since this model only has 24 total layers compared to GPT2-XL's 48) between the top-1 saturation layers, and the same top-1 class prediction in the final layer.
    \item For each image, as in all experiments conducted on this model we only consider the prediction at index 0 corresponding to the [CLS] token in the input.
    \item We used embeddings from 3 layers before and 3 layers after the saturation layer (instead of 5, again due to the smaller number of layers) resulting in 7 total forward passes.
\end{enumerate}

Figure \ref{fig:inject-result} shows that the results for this model follow a similar step function pattern to the ones for GPT2, where injecting embeddings from the top-1 saturation layer or one of the subsequent layers causes the model to "immediately" (at the injected layer) switch to the second task in a high percentage of cases, when compared to injecting embeddings from one of the layer before the top-1 saturation which almost never has the same effect.

\subsubsection{Intervention Procedure on Whisper} 
\label{sec:interv_whisper}
We made the following adjustments to run the procedure described in section \ref{interv_proc} on Whisper-large and 200 token pairs from randomly sampled 50 audios from the LibriSpeech  dataset:
\begin{enumerate}
    \item Since the average audio in LibriSpeech is 10 seconds long there are not enough tokens in one sample to find relevant pairs, so we wave the requirement for a pair to share context and only leave two conditions: a distance of at least 5 layers (instead of 10, since this model only has 32 total layers compared to GPT2-XL's 48) between the top-1 saturation layers, and the same top-1 prediction in the final layer.
    \item We used embeddings from 3 layers before and 3 layers after the saturation layer (instead of 5, again due to the smaller number of layers) resulting in 7 total forward passes.
\end{enumerate}

Figure \ref{fig:inject-result} shows that the results for this model follow a similar pattern to the other two models, even though the effect increases in the following layers after the saturation event.

\subsection{Additional Details for Token-Level Early Exit Measures Comparison} 
\label{sec:early-exit_detail}

\begin{figure}[ht]
    \begin{center}
    \includegraphics[width=0.5\linewidth]{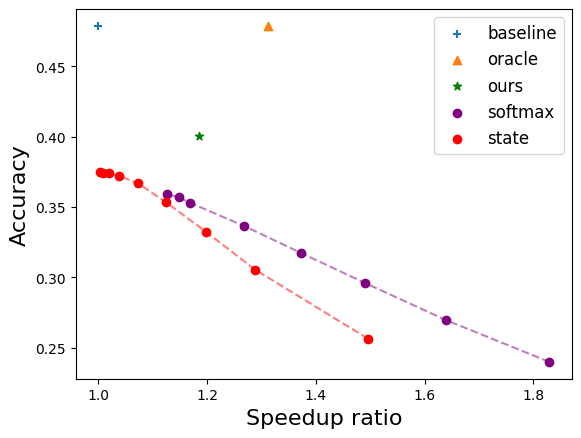} 
\end{center}
\caption{Performance-efficiency trade-off comparison of different confidence measures against a static baseline (where all layers are used for each token) and a local oracle measure (where the early exit is at the top-1 saturation layer). The graph shows softmax and state confidence measure results at different thresholds. Our method achieves the highest next-word prediction accuracy out of all early-exist methods while providing significant speedup compared to the baseline. }
\label {fig:early_exit}
\end{figure}

\begin{table}[ht]
\caption{Softmax Response accuracy \& speedup ratios at various confidence thresholds}
\begin{center}
\resizebox{\columnwidth}{!}{%
\begin{tabular}{lrrrrrrrr}
\toprule
\multicolumn{1}{c}{\bf }& \multicolumn{8}{c}{\bf Thresholds} \\
    & 0.4 & 0.5 & 0.6 & 0.7 & 0.8 & 0.9 & 0.92 & 0.94
\\\midrule
\bf Accuracy       & \(0.240\) & \(0.270\) & \(0.296\) & \(0.317\) & \(0.336\) & \(0.353\) & \(0.357\) & \(0.359\)\\
\bf Speedup Ratio              & \(1.830\) & \(1.640\) & \(1.491\) & \(1.373\)  & \(1.269\)& \(1.169\)&  \(1.148\)&  \(1.126\)\\
\\\bottomrule
\end{tabular}
}
\end{center}
\end{table}

\begin{table}[t!]
\caption{Hidden-sate saturation accuracy \& speedup ratios at various confidence thresholds}
\begin{center}
\resizebox{\columnwidth}{!}{%
\begin{tabular}{lrrrrrrrrr}
\toprule
\multicolumn{1}{c}{\bf }& \multicolumn{9}{c}{\bf Thresholds} \\
    & 0.986 & 0.988 & 0.989 & 0.99 & 0.991 & 0.992 & 0.993 & 0.994 & 0.995
\\\midrule
\bf Accuracy       & \(0.256\) & \(0.306\) & \(0.333\) & \(0.353\) & \(0.367\) & \(0.372\) & \(0.374\) & \( 0.374\) & \(0.375\)\\
\bf Speedup Ratio              & \(1.496\) & \(1.288\) & \(1.197\) & \( 1.125\)  & \(1.073\)& \(1.039\)&  \(1.019\)&  \(1.008\) &  \(1.003\)\\
\\\bottomrule
\end{tabular}
}
\end{center}
\end{table}

\end{document}